\documentclass[10pt,twocolumn,letterpaper]{article}

\usepackage[preprint]{cvpr}      % To produce the REVIEW version
\usepackage{graphicx}
\usepackage{amsmath}
\usepackage{amssymb}
\usepackage{booktabs}
\usepackage[table,xcdraw,dvipsnames]{xcolor}

\usepackage[pagebackref,breaklinks,colorlinks]{hyperref}
\usepackage{url}

\usepackage{enumitem} 
\usepackage{tabularx}
\usepackage{booktabs}
\usepackage{multirow} 
\usepackage{algorithm}
\usepackage{listings}

\usepackage{graphicx}
\usepackage{blindtext}
\usepackage{colortbl}
\usepackage{makecell}
\usepackage{balance}
\usepackage{adjustbox}
\usepackage{mathtools}
\usepackage{color}
\usepackage[accsupp]{axessibility}

\usepackage[capitalize]{cleveref}
\crefname{section}{Sec.}{Secs.}
\Crefname{section}{Section}{Sections}
\Crefname{table}{Table}{Tables}
\crefname{table}{Tab.}{Tabs.}

\def\cvprPaperID{2156} % *** Enter the CVPR Paper ID here
\def\confName{CVPR}
\def\confYear{2023}

\definecolor{mydarkblue}{rgb}{0,0.1,0.6}
\definecolor{Gray}{gray}{0.9}
\definecolor{Goldenrod}{RGB}{245,245,220} % 
\newcommand{\gain}[1]{\textcolor{Green}{(+{#1})}}% \textbf
\newcommand{\tablestyle}[2]{\setlength{\tabcolsep}{#1}\renewcommand{\arraystretch}{#2}\centering\small}

\hypersetup{
    colorlinks=true,
    citecolor=mydarkblue}

\begin{document}

\title{LUMix: Improving Mixup by Better Modelling Label Uncertainty}

\author{
	 Shuyang Sun$^{1}$\footnotemark[1]
	\;\;Jie-Neng Chen$^{2}$\footnotemark[1]
	\;\; Ruifei He$^3$
	\;\; Alan Yuille$^2$
	\;\; Philip Torr$^1$
	\;\; Song Bai$^4$ \\
	$^1$University of Oxford \;\; $^2$Johns Hopkins University \;\;  $^3$The University of Hong Kong \;\; $^4$ByteDance Inc. 
}
\twocolumn[{
\renewcommand\twocolumn[1][]{#1}
\maketitle
\begin{center}
    \centering
    \captionsetup{type=figure}
    \includegraphics[scale=1.5]{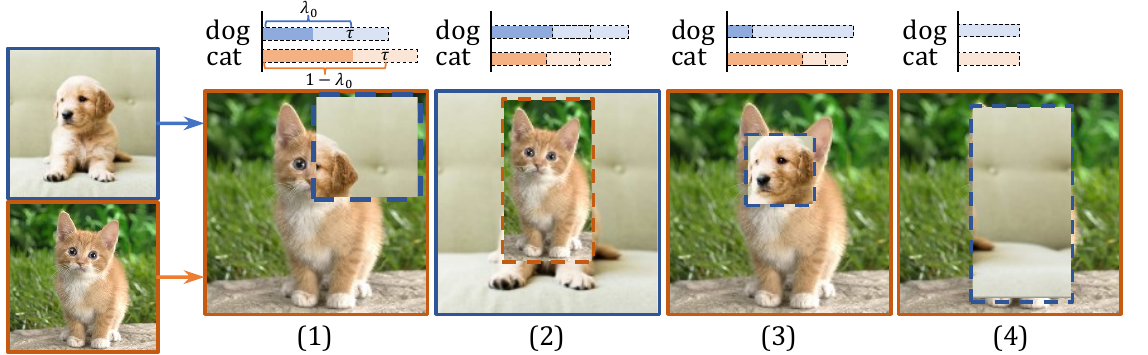}
    \caption{It is hard to estimate how much confidence should be given to each class in the above images. The label distribution is obviously not determined by the size of the crop. In this paper, we add perturbation $\tau$ to the label space to simulate such uncertainty. The original label weights $\lambda_0$ and $1-\lambda_0$ are calculated by the size of the crop. The label weight for the dog $\lambda$ can be calculated as $\lambda = \lambda_0 + \tau$. This simple modification help to model the label noise and provides a startling improvement to methods.
    Detailed implementation can be found in Algorithm \ref{alg:code}.}
    \label{fig:teaser}
\end{center}
}]

\renewcommand{\thefootnote}{\fnsymbol{footnote}}
\footnotetext[1]{These authors contributed equally to this work. 
Correspondence to Shuyang Sun (\url{kevinsun@robots.ox.ac.uk}) and Jie-Neng Chen (\url{jienengchen01@gmail.com}). }

\begin{abstract}
Modern deep networks can be better generalized when trained with noisy samples and regularization techniques. Mixup \cite{zhang2017mixup} and CutMix \cite{yun2019cutmix} have been proven to be effective for data augmentation to help avoid overfitting. Previous Mixup-based methods linearly combine images and labels to generate additional  training data. However, this is problematic if the object does not occupy the whole image as we demonstrate in Figure \ref{fig:teaser}. 
Correctly assigning the label weights is hard even for human beings and there is no clear criterion to measure it.
To tackle this problem, in this paper, we propose LUMix, which models such uncertainty by adding label perturbation during training. 
LUMix is simple as it can be implemented in just a few lines of code and can be universally applied to any deep networks \eg CNNs and Vision Transformers, with minimal computational cost.
Extensive experiments show that our LUMix can consistently boost the performance for networks with a wide range of diversity and capacity on ImageNet, \eg $+0.7\%$ for a small model DeiT-S and $+0.6\%$ for a large variant XCiT-L.
We also demonstrate that LUMix can lead to better robustness when evaluated on ImageNet-O and ImageNet-A. The source code can be found \href{https://github.com/kevin-ssy/LUMix}{here}.
\end{abstract}

\section{Introduction}
\label{sec:intro}
Modern deep networks \eg ResNet \cite{he2016deep}, ConvNeXt \cite{convnext}, Vision Transformer \cite{dosovitskiy2020image} are often over-parameterized which helps to achieve good performance. Recent advances in better training deep networks usually require massive data augmentation and regularization techniques to prevent the model from over-fitting. Among them, Mixup \cite{zhang2017mixup} and CutMix \cite{yun2019cutmix} have been found to be very helpful.

Mixup and CutMix first sample a random factor $\lambda \in [0, 1]$ that controls the proportion of the two training examples, then combines the original inputs to generate a new image with corresponding pseudo-labels via linear interpolation. 
However, we argue that such confidence re-weighting is not optimal. As shown in Figure \ref{fig:teaser}(1)(3), when combining different regions of the two images together, it is difficult to estimate the actual proportion with which the features of one image are perceivable with respect to the ones of the other image.

Meanwhile, as shown in Figure \ref{fig:teaser}(2)(4), the new crop may occlude the discriminative area on the canvas.
Labels in this case can also be misleading and bring false signals in the training process.
These problems have also been found in some other recent research \eg TransMix \cite{chen2021transmix}, PuzzleMix \cite{kim2020puzzle}, ResizeMix\cite{qin2020resizemix} and SaliencyMix \cite{uddin2020saliencymix}. However, these methods either exploit an extra model to rectify the false labels \cite{uddin2020saliencymix, chen2021transmix, kim2020puzzle} or simply avoid the uncertainty introduced in the mixing process \cite{qin2020resizemix}.

In this paper, unlike previous works that try to get rid of the label noise brought by the mixing process, we propose LUMix that will model the label noise. To this end, two terms are added on top of the original label mixture. One term is generated from the prediction of the deep network to judge whether there is a salient object in the input. 
The other is randomly sampled from a uniform distribution and is added to the label distribution to simulate the label uncertainty in the mixing process. Interestingly, we observe that such a simple addition works well and can perform on par with the recent state-of-the-art Mixup-based method like \cite{chen2021transmix}. We explore and reveal that the random noise introduced during the Mixup process can also be a causal factor to the improvement.

Experimental results demonstrate that such a simple tweak on data augmentation can lead to consistent improvement for many different models. For example, LUMix can help to improve the top-1 accuracy by $0.7\%$ for DeiT-S and $0.6\%$ for a large variant XCiT-L on ImageNet for image classification.
Furthermore, LUMix can also lead to better robustness on several challenging datasets \eg ImageNet-O and ImageNet-A when evaluated on three different benchmarks. With the better pre-training on ImageNet, we also find that the model pre-trained with LUMix can lead to better performance on Pascal VOC for semantic segmentation. LUMix can be implemented with just a few lines of code with almost no extra computation inducted. Meanwhile, unlike TransMix \cite{chen2021transmix} that can be only specifically applied to Vision Transformers, LUMix is also applicable to CNN architectures \eg ConvNext \cite{convnext} and ResNet \cite{resnet}.

Above all, the contribution of this paper can be summarized into two aspects:

\noindent(1) We propose a simple yet effective method that can further improve CutMix with minimal computational cost overhead. Extensive experiments are conducted to validated the efficacy of our method in terms of both accuracy and robustness for different tasks.

\noindent(2) 
Our method reveals that modelling label noise via perturbation can also be a feasible way to improve the network performance, which may inspire more future works to explore on this direction and help understand how Mixup-based methods work.

\section{Related Work}

\paragraph{Mixup and its variants.}
Mixup is a data augmentation technique that linearly interpolates in the input and label space with the same random proportion \cite{zhang2017mixup}. for better generalization, CutMix \cite{yun2019cutmix} mixes the inputs in a copy-paste approach so that the detailed texture of the original images can be kept.
Manifold-Mixup~\cite{verma2019manifold} further generalizes the Mixup method into the feature space for better performance. As extensions of the CutMix, Attentive-CutMix \cite{walawalkar2020attentive}, SnapMix\cite{huang2021snapmix}, Resize-CutMix \cite{qin2020resizemix} and saliency-CutMix \cite{uddin2020saliencymix} introduce another model to extract the foreground salient objects during the mixing process since the background part of the image is less relevant to the label. We argue that such an operation can somewhat narrow the augmentation space since the background regions can be also used as negative samples during training. MoEx\cite{moex} and PatchUp\cite{faramarzi2020patchup} propose to apply feature normalization to the image mixture, which also works well when incorporating with CutMix. StyleMix \cite{hong2021stylemix} applies techniques used in the field of style transfer to generalize the context of the mixed images.
Apart from improving the classification accuracy, there are also some papers that tend to improve the robustness of the network using Mixup related methods. For example, \cite{mixup_calibration} and \cite{on_mixup_training} found that Mixup can be used to improve the loss calibration and the robustness of the network. A concurrent work RecursiveMix \cite{yang2022recursivemix} iterates the mixing process for several times to further boost the performance, which in principle should be complementary to ours.
The recent TransMix \cite{chen2021transmix} also focuses on this problem but it chooses to manipulate the label space according to the attention map that the Vision Transformer \cite{dosovitskiy2020image} generates. Our work is most similar to TransMix since both TransMix and our method focus on label space manipulation. However, TransMix requires an additional attention map for re-weighting the labels so that it can be only compatible to Transformer-based networks with a class token at the end of the network. Instead, LUMix is model-agnostic and works well for both CNN and Transformer-based architectures.

\paragraph{Label smoothing.} Our proposed method is also related to label-smoothing  methods \cite{szegedy2016rethinking,delving_label_smoothing,muller2019does} since both LUMix and label-smoothing manipulate the label confidence. However, LUMix fundamentally differs from label-smoothing in the following aspects: (1) LUMix links each sample to a distribution in the label sapce, but label-smoothing changes the label weight to a fixed number. 
(2) LUMix only perturbs the classes to be mixed but label-smoothing changes weights for all classes. LUMix help to find a label distribution that works much better than label smoothing.
 (3) Label-smoothing is \textbf{complementary} with LUMix. We demonstrate in Section \ref{sec:exp} that baselines trained with label-smoothing can be further improved by LUMix by a clear margin.

\section{LUMix}
\subsection{A Recap on CutMix}
\label{sec:setup}
LUMix is built upon CutMix. Here we first recap the CutMix process for better understanding. Given two real data samples with the input images ${x}_{A}, {x}_{B}$ and their corresponding labels ${y}_{A}, {y}_{B}$ from the training dataset, the target of CutMix is to generate a new data sample $(\Tilde{{x}}, \Tilde{{y}})$ through mixing the two real samples together. The mixing process can be formulated as:
\begin{align}
    {\Tilde{x}} & = \mathbf{M} \odot {x}_{A} + (\mathbf{1}- \mathbf{M}) \odot {x}_{B},
    \label{eq:cutmix_x}
    \\
    {\Tilde{y}} & = \lambda {y}_A + (1-\lambda) {y}_B,
    \label{eq:cutmix_y}
\end{align}
where $\mathbf{M} \in \{0,1\}^{HW}$ is a binary mask with the size of the image $H\times W$,
 $\mathbf{1}$ is a map with same shape as $\mathbf{M}$ filled with ones, and $\odot$ is element-wise multiplication.  $\lambda \sim U(0, 1)$ is the proportion of ${y}_A$ in the mixed label. Here $U(0, 1)$ represents the uniform distribution from range [0, 1]. Sometimes the size of the crop from the image ${x}_{A}$ will exceed the boundary of the image ${x}_{B}$, in this case, we further clip the crop so that it can be well fit into ${x}_{B}$. The $\lambda$ should be re-calculated when the clipping operation happens. Note the $(\text{width}, \text{height})$ of the crop as $(r_w,r_h)$, then $\lambda = \frac{r_w r_h}{HW}$.

\subsection{LUMix}
\label{randmix}
As introduced in the above sections, the key difference between CutMix and LUMix is in the label space.
Here we add another random factor $\lambda_{r} \sim \text{Beta}(\alpha, \alpha)$ to the original proportion $\lambda_0$ sampled in CutMix:
\begin{align}
    \lambda = (1 - r_1 - r_2)\lambda_0 + r_1\lambda_r + r_2\lambda_s,
\end{align}
where $r_1, r_2$ are hyper-parameters that determines the ratio of $\lambda_r$ and $\lambda_s$ in the final $\lambda$. The introduction of $\lambda_r$ can be regarded as a way of simulating the human annotations as the human annotator is also uncertain about what confidence should be given to each class in the mixture. As for $\lambda_s$, that is calculated from the predicted score of the network, we assume the network produce scores with $C$ classes as ${p} \in \mathbb{R}^{C}$, then $\lambda_s$ is calculated as:
\begin{align}
    \hat p_i = \frac{e^{p_i}}{\sum_{j=1}^{C} e^{p_j}},
    \\
    \lambda_s = \frac{\hat{p}_A}{\hat{p}_A+\hat{p}_B},
\end{align}
where here $A, B$ indicate the indexes of the labeled classes of the input samples $A, B$. Since the score of the network can somewhat indicates the existence of the foreground object, the loss will tend to be lower if $\hat p_A < \lambda$ and vice versa.
Introducing $\lambda_s$ here is equivalent to do loss clipping solely on the target classes. In this way, the network will be less penalized if there is no valid object in the crop $A$ while performing CutMix data augmentation.

The introduction of $\lambda_s$ requires the network prediction to be reliable, however, some recent papers \cite{on_mixup_training, zhang2017mixup} found that deep networks tend to be under-confident when the $\lambda$ is sampled from a Beta distribution $\text{Beta}(\alpha, \alpha)$ with $\alpha > 0.4$. All experiments in this paper are also conducted with Mixup with $\alpha = 0.8$.
To alleviate this problem, we add a new regularization term $\mathcal{R}$ that helps to better calibrate the network prediction:
\begin{equation}
    \mathcal{L} = \mathcal{L}_0 + \eta \mathcal{R},
\end{equation}
where $\mathcal{L}_0$ is the value of loss calculated using SoftMax or binary cross entropy with softened labels. $\eta$ is a hyper-parameter that controls the influence of the regularization factor $\mathcal{R}$. 
\begin{align}
    \mathcal{R} &= \sum_{k=1}^{C} y_k\text{max}(0, b_k - \hat{p}_k),
    \\
    {b} &= {y}_A \land {y}_B,
\end{align}
where ${b} \in \{0, 1\}^{C}$ is a binary vector. Here $\mathcal{R}$ is similar to the Hinge loss that only penalizes on the positive classes.

%##################################################################################################
\begin{algorithm}[t]
\caption{\small{Pseudo-code of RandMix in PyTorch style.}}
\label{alg:code}

\definecolor{codeblue}{rgb}{0.25,0.5,0.5}
\lstset{
  backgroundcolor=\color{white},
  basicstyle=\fontsize{7.2pt}{7.2pt}\ttfamily\selectfont,
  columns=fullflexible,
  breaklines=true,
  captionpos=b,
  commentstyle=\fontsize{7.2pt}{7.2pt}\color{codeblue},
  keywordstyle=\fontsize{7.2pt}{7.2pt},
%  frame=tb,
}
\begin{lstlisting}[language=python]
# B: number of images in a batch (batch size).
# C: number of classes
# logits: output of the network
# alpha: the hyper-parameter for Beta distribution
# y1, y2: labels for different samples
# r_s, r_r: ratios for lam_s and lam_r
# ce_loss: the default softmax cross entropy loss

def randmix(logits, y1, y2):
    # y1, y2: [B, C]
    scores = F.softmax(logits, dim=-1)
    prob1, prob2 = scores[y1 > y1.min()], scores[y2 > y2.min()]
    # lam_s, lam_r, lam: [B]
    lam_s = prob2 / (prob1 + prob2)
    # samples lam_r for B times from a beta distribution.
    lam_r = beta(alpha, alpha, B)
    lam = lam0 * (1 - r_s - r_r) + lam_s * r_s + lam_r * r_r
    y = (1-lam) * y1 + lam * y2
    # generate the binarized label b
    b[y1 > y1.min()] = 1
    b[y2 > y2.min()] = 1
    loss = ce_loss(logits, y) + y * max(0, b - scores)
\end{lstlisting}
\end{algorithm}
%##################################################################################################
\subsection{Pseudo-code in PyTorch style}
The pseudo-code of LUMix is provided in Algorithm \ref{alg:code}. We demonstrate that our method can be implemented neatly with about ten lines of code. We also emphasize that all operations used in the code are fast, simple and supported by current popular deep learning libraries \eg PyTorch, TensorFlow and Jax.

\section{Experiments}
\label{sec:exp}
\subsection{ImageNet Classification}
\label{sec:cls}
\paragraph{Implementation details}
\label{sec:impl}
We benchmark all the models reported in this paper on ImageNet-1k \cite{deng2009imagenet}.
 There are 1.28M images for training and 50k images for validation in ImageNet-1k. For all the models trained on ImageNet, we report their top-1 accuracy on the validation set for fair comparison. We build all the training pipelines under the implementation of Timm~\cite{rw2019timm} library in PyTorch. Unless specified otherwise, for all models reported in this paper, we use the same set of hyper-parameters and training regime as RandAug~\cite{cubuk2020randaugment}, Stochastic Depth (Drop Path)~\cite{huang2016deep}, Mixup~\cite{zhang2017mixup} and CutMix \cite{yun2019cutmix}. The learning rates for all models reported in this paper are set to 0.001 and are linearly warmed up from $1\times 10^{-6}$ for 20 epochs, except DeiT, which is only warmed up for 5 epochs. All models are trained and evaluated using a resolution of $224\times224$ on 8 A100 GPUs. For evaluation, all models are center-cropped with a percentage $0.875$ except the XCiT ($1.0$) as reported in their original paper.

%##################################################################################################
\begin{table}[t]
\tablestyle{1.8pt}{1.3}
\centering
\begin{tabular}{cccccc}
% \hline
\toprule[1.5pt]
Models            & \#Params & FLOPs & Epochs & \begin{tabular}[c]{@{}c@{}}Top-1 \\ (\%) \end{tabular}& \begin{tabular}[c]{@{}c@{}} +LUMix \\ Top-1 (\%) \end{tabular}      \\ \hline
ConvNeXt-T~\cite{convnext}           & 29M    & 4.5G  & 300 & 82.1  & \cellcolor{Gray!36} \textbf{82.5}  \\
Swin-T~\cite{liu2021swin}            & 28M    & 4.5G  & 300 & 81.3  & \cellcolor{Gray!36} \textbf{81.7}  \\
DeiT-S~\cite{touvron2021training}    & 22M    & 4.7G  & 300 & 79.8  & \cellcolor{Gray!36} \textbf{{80.6}}  \\
XCiT-S~\cite{el2021xcit}             & 26M    & 4.8G  & 400 & 82.0  & \cellcolor{Gray!36} \textbf{{82.3}} \\
\hline
ConvNeXt-S~\cite{convnext}           & 50M    & 8.7G  & 300 & 83.1  & \cellcolor{Gray!36} \textbf{83.3}  \\
Swin-S~\cite{liu2021swin}            & 50M    & 8.7G  & 300 & 83.0  & \cellcolor{Gray!36} \textbf{{83.1}}  \\
\hline
ConvNeXt-B~\cite{convnext}           & 89M    & 15.4G & 300 & 83.8  & \cellcolor{Gray!36} \textbf{84.1}  \\
Swin-B~\cite{liu2021swin}            & 88M    & 15.4G & 300 & 83.5  & \cellcolor{Gray!36} \textbf{{83.6}}  \\
XCiT-M~\cite{el2021xcit}             & 84M    & 16.2G & 400 & 82.7  & \cellcolor{Gray!36} \textbf{83.2}  \\
DeiT-B~\cite{touvron2021training}    & 87M    & 17.6G & 300 & 81.8  & \cellcolor{Gray!36} \textbf{82.2}  \\
\hline
XCiT-L~\cite{el2021xcit}             & 189M   & 36.1G & 400 & 82.9  & \cellcolor{Gray!36} \textbf{{83.6}}  \\
\toprule[1.5pt]
\end{tabular}
\caption{LUMix can steadily boost the a wide range of model variants \eg DeiT, Swin Transformer, XCiT and also a CNN-based network ConvNeXt on ImageNet-1k for image classification. Note that all the baselines have been already carefully tuned with extensive augmentation and regularization techniques \eg Mixup \cite{zhang2017mixup}, CutMix\cite{yun2019cutmix}, RandAug\cite{cubuk2020randaugment}, DropPath\cite{droppath} \etc.}
\label{tab:cls}
\end{table}
\paragraph{Experimental results}
We show that LUMix can help to improve all the listed models in Table \ref{tab:cls}. These models are of a wide range of capacity in terms of both the number of parameters (from 22M to 189M) and FLOPs (from 4.5G to 36.1G). We note that all these baselines are strong since they are carefully tuned with massive data augmentation and regularization techniques. Therefore, the improvements that LUMix achieves on these models are remarkable. Concretely, for small models like Deit-S, LUMix can achieve a $0.7\%$ gain on ImageNet. 
As for the large variant like XCiT-L, LUMix can also lead to a $0.6\%$ gain. We emphasize that it is surprising that such a tiny change on data augmentation with minimal cost overhead can notably lift the overall performance of a wide range of model architectures.

\begin{table}[t]
\begin{minipage}{0.46\textwidth}
\tablestyle{6.6pt}{1.2}
\begin{tabular}{ccccc}
\toprule[1.5pt]
Model & $\lambda_s$ & $\lambda_r$ & $\mathcal{R}$ &  \makecell[c]{Top-1 \\Acc} \\
\hline
\multirow{5}{*}{DeiT-S} &  & & &  79.8\% \\
 & \checkmark &  & &  80.3\% \\
 &  & \checkmark & &  80.3\% \\
  & \checkmark & \checkmark & &  80.4\% \\
  & \checkmark & \checkmark & \checkmark &  80.5\% \\
\toprule[1.5pt]
\end{tabular}
% \vspace{3px}
\caption{ \textbf{Component analysis on $\lambda_s$, $\lambda_r$ and $\mathcal{R}$.} We observe that the randomization in the label space results in the majority of the total improvement. }
\label{tab:abl}
\end{minipage}
\hfill
\begin{minipage}{0.46\textwidth}
% \begin{table}[t]
\tablestyle{6.6pt}{1}
\begin{tabular}{ccccc}
\toprule[1.5pt]
Model & $r_1$ & $r_2$ & \makecell[c]{Top-1 \\Acc} \\
\hline
\multirow{6}{*}{DeiT-S} & 0.0 & 0.0 & 79.8\% \\
 & 1.0 & 0.0 &  79.7\% \\
 & 0.5 & 0.0 &  80.3\% \\
 & \cellcolor{Gray!36}0.4 & \cellcolor{Gray!36}0.1 &  \cellcolor{Gray!36}80.4\% \\
 & 0.3 & 0.2 &  80.3\% \\
 & 0.0 & 0.5 &  80.3\% \\
\toprule[1.5pt]
\end{tabular}
% \vspace{3pt}
\caption{{Ablation study on different random ratios $r_1$, $r_2$ with backbone Deit-S.} $r_1 = 1$ means that the target confidences are completely random.}
\label{tab:ratio}
% \end{table}
\end{minipage}

\end{table}
\subsection{Ablation studies}
\paragraph{The effect of each component in LUMix}
LUMix consists of three different components, $\lambda_r$, $\lambda_s$ and the regularization $\mathcal{R}$. We show the relative effects of applying these factors for DeiT-S in Table \ref{tab:abl}. From Table \ref{tab:abl} we can find that either $\lambda_r$ or $\lambda_s$ can lead to remarkable improvement (+$0.6\%$) on ImageNet in terms of top-1 accuracy. Applying an auxiliary loss $\mathcal{R}$ for such heavily regularized network can also lead to {$0.1\%$} gain for top-1 accuracy. Apart from producing more accurate predictions, in the following sections we further show that by training together with LUMix, the prediction of the network can be better calibrated so that the network can be more robust.

\begin{table*}[t]
\begin{minipage}{0.48\textwidth}
% \begin{table}[t]
\tablestyle{1.6pt}{1.2}
\begin{tabular}{cccccc}
\toprule[1.5pt]
Backbone & Decoder & \begin{tabular}[c]{@{}c@{}}LUMix- \\ pretrained\end{tabular} & mAcc & mIoU  & \multicolumn{1}{c}{\makecell[c]{mIoU \\(MS)}} \\ \hline
ResNet101 & Deeplabv3+ & & 57.4 & 47.3 & 48.5 \\ \hline
\multirow{2}{*}{DeiT-S} 
 &   \multirow{2}{*}{Segmenter} & & 60.4 &  49.7 & 50.5 \\
 & & \cellcolor{Gray!36}\checkmark & \cellcolor{Gray!36}\textbf{60.9} & \cellcolor{Gray!36}\textbf{50.3} & \cellcolor{Gray!36}\textbf{50.7} \\ 
\toprule[1.5pt]
\end{tabular}
\caption{Performance (mAcc, mIoU) when transferring DeiT-S to semantic segmentation. The model pre-trained with LUMix on ImageNet-1k can perform slightly better than the baseline. (MS) denotes multi-scale testing.}
\label{tab:seg}
% \end{table}
\end{minipage}
\hfill
\begin{minipage}{0.48\textwidth}
% \begin{table}[t]
\tablestyle{1.6pt}{1.6}
\begin{tabular}{ccc}
% \hline
\toprule[1.5pt]
 & \makecell[c]{Seg JI (\%)} & \makecell[c]{Loc mIoU (\%)} \\ \hline
DeiT-S & 29.2 & 34.9 \\
\rowcolor{Gray!36}
LUMix-DeiT-S & \textbf{29.5} & \textbf{41.8} \\ 
\toprule[1.5pt]
\end{tabular}
\caption{Quantitative evaluation of weakly supervised segmentation and location. Seg JI denotes the Jaccard index for WSAS on Pascal VOC. Loc mIoU denotes the bounding box mIoU for WOSL on ImageNet-1k.} % -1k val. 
% \end{table}
\end{minipage}
\end{table*}
\paragraph{Ablation of different random ratios} Table \ref{tab:ratio} exhibits the top-1 accuracy of Deit-S when applying different ratios to random factors $\lambda_r$ and $\lambda_s$ in LUMix. We note that when $r_1 + r_2 = 1$, then the label weights for the positive classes are completely randomly re-assigned. It is surprising to find that the network only degrades by just $0.1\%$ when the label confidence were completely randomized. Adding some randomness to the label space can significantly improve the final performance but adjusting the ratio $r_{1}, r_2$ can only lead to a incremental $0.1\%$ gain.

\subsection{Transferability to Downstream Tasks}
\label{sec:tra}
In this section, we transfer the model pre-trained on ImageNet and evaluate them on downstream tasks including semantic segmentation and weakly supervised object localization.
We validate that the improvement on ImageNet of LUMix can be further transferred to other downstream tasks.

\paragraph{Semantic Segmentation}
In the semantic segmentation task, the feature map from encoder (\eg Vision Transformer) is processed with a decoder to  generate the segmentation map $s\in \mathbb{R}^{H \times W \times N}$ where N is the number of semantic classes. We keep the DeiT-S as the segmentation encoder and adopt the convolution-free Segmenter~\cite{strudel2021} as the semantic segmentation decoder. 
The Segmenter~\cite{strudel2021} decoder is a Transformer-based decoder, namely Mask Transformer introduced in ~\cite{strudel2021, wang2021max}.

All the results on semantic segmentation are evaluated on the Pascal Context ~\cite{mottaghi_cvpr14} dataset. We follow the commonly used Intersection over Union (mIoU) metric averaged over all classes to compare between different methods. Pascal Context has 4998 images for training and it is divided into 59 semantic classes plus a background class. There are 5105 images for validation. 
We follow the training regime proposed in ~\cite{mottaghi_cvpr14} and implement everything using MMSegmentation \cite{mmseg2020}. 
To show that our baseline is strong, we further list the result of ResNet101-Deeplabv3+~\cite{chen2017deeplab, chen2018encoder} in Table \ref{tab:seg} as a reference.
According to Table~\ref{tab:seg}, LUMix improve over the vanilla pre-trained baselines by 0.6\% mAcc and 0.9\% mIoU respectively. There are consistent improvements on multi-scale testing.

\paragraph{Weakly Supervised Object Segmentation and Localization}
Apart from the semantic segmentation, we further evaluate the model trained with our method on two weakly supervised tasks (1) {\textit{Weakly Supervised Automatic Segmentation (WSAS)}} on Pascal VOC 2012 benchmark~\cite{pascal-voc-2012} and (2) {\textit{Weakly Supervised Object Localization (WOSL)}} on ImageNet-1k validation set~\cite{russakovsky2015imagenet} to show the efficacy of it. We note that the bounding box labels in these two datasets are only available for evaluation. 
We train the model on ImageNet-1k training set then follow \cite{naseer2021intriguing} to evaluate the quality of attention matrix by thresholding it into a binary attention mask.
 For WSAS, we measure the performance using the Jaccard similarity between ground truth on PASCAL-VOC12 and binary attention masks generated from DeiT-S.
As for WOSL, since the attention map itself can be used to localize the object, there is no need for us to use CAM-based methods. Instead, we can directly generate a bounding box from the binary attention masks as the prediction, then compare it with the ground truth.
The attention masks generated from LUMix-DeiT-S or vanilla DeiT-S are compared with ground-truth on these two benchmarks. The evaluated scores can quantitatively help us to understand if LUMix has a positive effect on the quality of attention map.

\subsection{Robustness Analysis}
\label{sec:rob}
\noindent\textbf{Robustness to Occlusion}
We follow \cite{naseer2021intriguing} to study the robustness of ViTs under occluded scenarios, where some or most of the image content is missing. To be specific, Vision Transformers divide the input image into 14x14 spatial grid \ie an image with resolution 224$\times$224 will go into 196 patches. Here we conduct Patch Dropping as a way attacking the input, which zero-masks the original image patches. 
As an example, dropping 100 patches from the input actually loses about a half of the image content.
Following \cite{naseer2021intriguing}, we evaluate the classification accuracy on ImageNet-1k validation set with the following different settings. (1) \emph{Random Patch Dropping} randomly selects M patches of the input image and drops them. 
(2) \emph{Salient (foreground) Patch Dropping} drops patches  in the highly salient regions to investigate the robustness of the network. The salient patches are sampled from the attention map learnt from DINO \cite{caron2021emerging}.
(3) \emph{Non-salient (background) Patch Dropping} drops patches in non-salient regions of the image. The non-salient regions selection follows the same approach as in (2). As shown in Table \ref{tab:occlude3}, DeiT-S with LUMix is more robust against occlusion than vanilla DeiT-S, especially for background Patch Dropping.  % ~\ref{tab:occlude1}, \ref{tab:occlude2},

\paragraph{Robustness to Spatial Shuffling}
We remove the structural information within images (spatial relationships) by defining a shuffling operation on input image patches, such that we can study the model’s sensitivity to the spatial structure by shuffling on input image patches. To be specific, we shuffle the image patches randomly with different grid sizes following~\cite{naseer2021intriguing}. Note that a shuffle grid size of 1 means no shuffle, and a shuffle grid size of 196 means all patch tokens are shuffled. 
Table~\ref{tab:shuffle} shows the consistent improvements over the baseline, and the accuracy averaged on all shuffled grid sizes for LUMix-DeiT-S and DeiT-S are 
62.3\% and 61.8\% respectively.  The result indicates that LUMix enables Transformers to rely less on positional embedding to preserve the most informative context for classification.

\begin{table}[t]
    \tablestyle{0.6pt}{1.3}
% \scalebox{0.9}{
\begin{tabular}{lccccl}
\toprule[1.5pt]
Method & Backbone & \#Params & Speed (fps) & \makecell[c]{Model\\Agnostic?}& \makecell[c]{Top-1 Acc\\(\%)} \\
\hline
Baseline & \multirow{6}{*}{DeiT-S} &  22M & 322 & - &  78.6 \\
CutMix & &  22M & 322      & \checkmark &  79.8 \gain{1.2} \\
SaliencyMix &  &  22M & 314 & \checkmark &  79.2 \gain{0.6} \\
Puzzle-Mix &  &  22M & 139 & \checkmark &  79.8 \gain{1.2} \\
TransMix & &  22M & 322 & $\times$ &  \textbf{80.7} \gain{2.1} \\
\rowcolor{Gray!36}
LUMix & &  22M & 322 & \checkmark & \textbf{80.6} \gain{2.0} \\
\hline
Baseline & \multirow{5}{*}{ResNet-50} &  22M & 464 & - &  76.3 \\
CutMix &  &  22M & 464      & \checkmark &  78.6 \gain{2.3} \\
SaliencyMix & &  22M & 433 & \checkmark &  78.7 \gain{2.4} \\
Puzzle-Mix & &  22M & 176 & \checkmark &  78.8 \gain{2.5} \\
\rowcolor{Gray!36}
LUMix & &  22M & 464 & \checkmark & \textbf{79.1} \gain{2.8} \\
\toprule[1.5pt]
\end{tabular}
\caption{\textbf{Comprehensive comparison with state-of-the-art Mixup variants on ImageNet-1k}. All listed models are built upon DeiT training recipe for fair comparison. Here TransMix is only compatible with Transformer-based network that is with a class-token at the end of the network.}
\label{tab:other_mixup}
\end{table}

\paragraph{Natural Adversarial Example}
ImageNet-Adversarial (ImageNet-A) dataset~\cite{hendrycks2021natural} collects 7500 unmodified, naturally occurring but ``hard" images. These examples are gathered from some challenging scenarios (\eg fog scene, noise, and occlusion) which will significantly degrade the recognition performance for the machine learning models. The metric for assessing classifiers' robustness to adversarially filtered examples includes the top-1 accuracy, Calibration Error (CalibError)~\cite{kumar2019verified, hendrycks2021natural}, and Area Under the Response Rate Accuracy Curve (AURRA). Calibration Error measures how classifiers can reliably forecast their accuracy. AURRA is an uncertainty estimation metric introduced in \cite{hendrycks2021natural}. As shown in Table~\ref{tab:ood}, DeiT-S trained under LUMix is superior to vanilla DeiT-S on all metrics.

\paragraph{Out-of-distribution Detection}
The ImageNet-O~\cite{hendrycks2021natural} is an  out-of-distribution detection dataset, which adversarially collects 2000 images from a distribution unlike the ImageNet training distribution. The anomalies of unforeseen classes should result in low-confidence predictions. The metric is the area under the precision-recall curve (AUPR)~\cite{hendrycks2021natural}. Table~\ref{tab:ood} indicates that DeiT-S with LUMix outperform DeiT-S by 0.4\% AUPR.

\begin{table}[t]
\begin{minipage}{0.48\textwidth}
\centering
\tablestyle{1.8pt}{1.2}
\begin{tabular}{l|ccccccc}
\toprule[1.5pt]
Grid size     & 1    & 4  & 8    & 16   & 32 & 64  & Avg  \\
\hline
DeiT-S        & 79.9 & 75.6 & 73.3 & 69.2 & 59.8 & 46.1 & 67.3 \\
\rowcolor{Gray!36}
\makecell[l]{DeiT-S\\+LUMix}   & 80.5 & 76.3 & 73.6 & 69.9 & 60.3 & 46.3 & \textbf{67.8 } \\
\toprule[1.5pt]
\end{tabular}
% \vspace{-6pt}
\caption{Model's robustness to spatial structure shuffling.}
\label{tab:shuffle}
\vspace{6px}
\tablestyle{1.6pt}{1.6}
\begin{tabular}{cccc}
\toprule[1.5pt]
Type & Metric & DeiT-S & \makecell[c]{LUMix-\\DeiT-S}\\
\hline
\multirow{3}{*}{\makecell[c]{Natural\\Adversarial\\Example}} & Top1-Acc & 19.1 & \cellcolor{Gray!36} \textbf{19.9} \\
 & Calib-Error$\downarrow$ & 32.0 & \cellcolor{Gray!36} \textbf{31.5} \\
 & AURRA & 23.8 & \cellcolor{Gray!36}  \textbf{27.4} \\
 Out-of-Dist & AUPR & 20.9 & \cellcolor{Gray!36}  \textbf{21.3} \\
\toprule[1.5pt]
\end{tabular}
% \vspace{-6pt}
\caption{Model's robustness against natural adversarial examples on ImageNet-A and out-of-distribution examples on ImageNet-O.}
\label{tab:ood}
\end{minipage}
\end{table}

\begin{table*}[t]
\tablestyle{6.6pt}{1.2}
\begin{tabular}{lccccccccccc}
\toprule[1.5pt]
Information Loss (\%)   & 0    & 10   & 20   & 30   & 40   & 50   & 60   & 70   & 80   & 90   & Avg  \\ \hline
DeiT-S Acc (\%)         & 79.8 & 76.7 & 71.9 & 65.9 & 59.1 & 50.5 & 40.9 & 29.5 & 17.4 & 5.7  & 49.75 \\
\rowcolor{Gray!36}
LUMix-DeiT-S Acc (\%) & 80.6 & 77.5 & 73.2 & 68.2 & 62.0 & 53.9 & 44.8 & 34.4 & 23.0 & 11.7 & \textbf{52.92} \\ 
\toprule[1.5pt]
\end{tabular}
\caption{Top-1 accuracy on ImageNet-val to occlusion (\ie Non-Saliency Patch Dropping) for DeiT-S trained \textit{w/} and \textit{w/o} LUMix.}
\label{tab:occlude3}
\end{table*}
\subsection{Comparison with Other Mixup Variants}
\label{sec:sota}
In this section, we compare our method with other state-of-the-art Mixup-based methods on ImageNet-1k. We re-implement all the methods reported in Table \ref{tab:other_mixup} and train a DeiT-S under the same training regime, except for the application of these Mixup-based methods for fair comparison. The baseline reported in Table \ref{tab:other_mixup} is trained with all other data augmentation and regularization techniques we described in Section \ref{sec:impl} except for CutMix. We measure the training speed (throughput) of each method on a NVIDIA V100 GPU with a batch size of 128.

Table~\ref{tab:other_mixup} shows that LUMix can outperform all the other Mixup methods except for TransMix. However, TransMix is only compatible with Transformer-based networks while our LUMix is \textbf{model-agnostic} and can be well generalized onto CNN architectures like ConvNeXt~\cite{convnext} and ResNet \cite{resnet}. We emphasize that this is crucial since most networks, even for Transformer-based networks \eg Swin Transformer, do not contain a class token to generate the attention map that TransMix requires.
When compared with other model-agnostic Mixup-based methods, \eg Attentive-CutMix, SaliencyMix and Puzzle-Mix, LUMix can be superior to all these methods in terms of accuracy-speed trade-off. Concretely, LUMix can outperform Puzzle-Mix and SaliencyMix by $0.7\%$ and $1.3\%$ respectively in terms of the top-1 accuracy on ImageNet. We also compare LUMix and other state-of-the-art Mixup methods based on ResNet in Table \ref{tab:other_mixup}, from which we can conclude that LUMix can outperform these methods in terms of both accuracy and speed.

\begin{table}[t]
\begin{minipage}{0.48\textwidth}
\tablestyle{6.6pt}{1.1}
\begin{tabular}{ccccc}
\toprule[1.5pt]
Model & \makecell[c]{Distribution \\for $\lambda_r$} & \makecell[c]{Top-1 \\Acc} \\
\hline
\multirow{7}{*}{DeiT-S} & - & 79.8\% \\
 & $\mathcal{N}(0, 1)$ & 80.4\% \\
 & $\mathcal{N}(0, 5)$ & 80.4\% \\
 & Beta(0.8, 0.8) & 80.3\% \\
 & \cellcolor{Gray!36}Beta(2.0, 2.0) & \cellcolor{Gray!36}\textbf{80.6\%} \\ 
 & Beta(3.0, 3.0) & 80.4\% \\
\toprule[1.5pt]
\end{tabular}
\vspace{-6pt}
\caption{{Ablation on different random ratios $r_1$, $r_2$ with backbone Deit-S.} $r_1 = 1$ means that the target confidences are completely random.}
\label{tab:alpha}
\end{minipage}
\end{table}

\section{Discussion and Analysis}
Assume we have a training dataset $\mathcal{D} = \{x_{i}, y_{i}\}^{n}_{i=1}$ where $(x_{i}, y_{i}) \sim P$ for all data samples in $\mathcal{D}$ with $n$ labeled samples. $P$ is a unknown joint distribution between the input and the target space. We can use a empirical Dirac delta distribution $\delta$ centered at $(x_i,y_i)$ to approximate $P$ as follows:
\begin{align}
P_\delta (x,y) = \frac{1}{n} \sum_i^n \delta(x=x_i, y=y_i).
\end{align}
However, there can be lots of possible estimates for $P$ and $P_\delta$ is just one of then. To generalize $P_\delta$, we can replace the Dirac delta distribution $\delta$ with a vicinity distribution $\nu$:
\begin{align}
P_\nu(\tilde{x},\tilde{y}) = \frac{1}{n} \sum_i^n \nu(\tilde{x},\tilde{y}|x_i,y_i).
\end{align}
Mixup \cite{zhang2017mixup} essentially creates a new vicinity distribution $\mu(\tilde{x}, \tilde{y} | x_i, y_i)$ that synthesizes new data sample pairs $\tilde{x}, \tilde{y}$ through linear interpolation:
\begin{multline}
    \mu(\tilde{x}, \tilde{y} | x_i, y_i) = \frac{1}{n} \sum_j^n\mathbb{E}_{\lambda} \bigl[ \delta(\tilde{x}
   = \lambda \cdot x_i + (1-\lambda) \cdot x_j,\\ \tilde{y} = \lambda \cdot y_i +
   (1-\lambda) \cdot y_j) \bigr],
\end{multline}

\begin{table}[t]
    \tablestyle{3.6pt}{1.3}
\begin{tabular}{cccc}
\toprule[1.5pt]
Model & Loss & \textit{w/} LUMix? &  \makecell[c]{Top-1 \\Acc} \\
\hline
\multirow{4}{*}{DeiT-S} & SCE & &  79.8\% \\
 & \cellcolor{Gray!36}SCE & \cellcolor{Gray!36}\checkmark & \cellcolor{Gray!36}80.6\% \\
 & BCE & &  79.9\% \\
 & BCE & \checkmark &  79.6\% \\
\toprule[1.5pt]
\end{tabular}
% \vspace{3px}
    \caption{Effect of applying different loss functions to DeiT-S and LUMix. SCE and BCE are abbreviations for SoftMax and Binary Cross Entropy.}
    \label{tab:bce}
\end{table}
The success of Mixup proves that such linear inductive bias can help to generalize the training space \cite{zhang2017mixup}. Unlike Mixup that equally assigns the $\lambda$ to all pixels on the input images, CutMix uses a binary map to mix the two images together. 
However, as discussed in Section \ref{sec:intro} and some recent papers \cite{chen2021transmix, kim2020puzzle, uddin2020saliencymix}, we argue that this may cause problems for CutMix since {not all pixels are created equal}. 
The binary masking defined in Eq. \eqref{eq:cutmix_x} can make the semantic meaning of the generated crop completely irrelevant to the label.
This indicates that for CutMix, some of the distriminative contents of the two inputs may be filtered out so that the label can become misleading. Therefore, add a random factor to the label space can help to bridge the gap between the vicinity distribution and the true distribution. We can denote the new vicinity distribution $\mu'(\tilde{x}, \tilde{y} | x_i, y_i)$ by:
\begin{multline}
    \mu'(\tilde{x}, \tilde{y} | x_i, y_i) = \frac{1}{n} \sum_j^n\mathbb{E}_{\lambda} \bigl[ \delta(\tilde{x} = \mathbf{M} \odot x_i + \\(\mathbf{1}-\mathbf{M}) \odot x_j,
    \tilde{y} = (\lambda+r) y_i + (1-\lambda-r)y_j) \bigr],
\end{multline}

where $\tau \sim \mathcal{H}$ represents all random factors inducted into the vicinity distribution, which in practice includes the $\lambda_r$ and $\lambda_s$ defined in previous sections. Here $\mathcal{H}$ is a distribution that simulates the random noise during the image mixing process.
The introduction of such randomness into the label space validates our observation that the label space for Mixup-based methods should consider some extent of ambiguity as it is even hard for human beings to estimation how uncertain they are about their uncertainty.

\begin{table}[t]
    \tablestyle{3.6pt}{1.3}
    \begin{tabular}{ccccc}
        \toprule[1.5pt]
        Model & \textit{w/} RS? & \textit{w/} LM? & \textit{w/} R$\lambda$? &  \makecell[c]{Top-1 \\Acc} \\
        \hline
        \multirow{4}{*}{DeiT-S} & \checkmark & & & 79.8\% \\
         & \checkmark & \checkmark & & 80.4\% \\
         & & \checkmark & & 80.6\% \\
         & & & \checkmark & 79.1\% \\
    \toprule[1.5pt]
    \end{tabular}
    \caption{The effect of applying LUMix (LM), Random Shuffling (RS) and Random $\lambda$ (R$\lambda$).}
    \label{tab:randomness}
\end{table}

\paragraph{What distribution shall we choose to sample $\lambda_r$?}
As discussed above, the introduction of the random factor in the label space can help to bridge the gap between the original vicinity distribution to the true distribution. However, there is still a problem remaining, what distribution of $\lambda_r$ can help the most? Or, how can we sample $\lambda_r$ to make the vicinity distribution closer to the true distribution? In this paper, we try to sample $\lambda_r$ from two different distribution, the Gaussian distribution $\mathcal{N}(0, 1)$ and the Beta distribution $\text{Beta}(\alpha, \alpha)$. As shown in Table \ref{tab:alpha}, we investigate the performance of LUMix when sampling $\lambda_r$ under different distributions. Note that the Beta distribution $\text{Beta}(1, 1)$  is equivalent to the unit uniform distribution $U(0, 1)$. From Table \ref{tab:alpha}, we conclude that our method works the best in terms of top-1 accuracy, when $\lambda_r$ is sampled from a Beta distribution with $\alpha=2$. In the future, it is interesting to investigate how to \textit{learn a distribution} to sample $\lambda_r$ that can further help the vicinity distribution to achieve better performance.

\paragraph{Compatibility with binary cross entropy loss.} As CutMix introduces a multi-class classification problem, it may be more suitable to use the Binary Cross Entropy (BCE) loss to supervise the network. The BCE loss does not force each class to be mutually exclusive so we do not need to re-weight the target confidence for each class, which avoids the problem about how to estimate the uncertainty for each class.
We show the Top-1 accuracy of DeiT-S trained with BCE loss in {Table \ref{tab:bce}}, but find that it can only achieve marginal gain compared to the baseline result that is trained under SoftMax cross entropy. When combined together with LUMix that further softens the target confidence of the training labels, the model performs even worse than the baseline result. We suggest that unlike SoftMax cross entropy, in binary cross entropy, adding relaxation to the positive classes of the targets can always lead to weaker penalization, since the relaxed target confidences are always lower than their original values.

\paragraph{Adding more randomness to the mixing process.} In the previous sections, we have demonstrated that adding random noise into the label space can help improve the model performance. We wonder whether adding more randomness into either the input or the label space can bring further improvement. Table \ref{tab:randomness} exhibits that when applying random shuffling and random $\lambda$ on input images. Here the random shuffling first divides the image into a set of 16$\times$16 patches then applying random shuffling for the mixed image. Random $\lambda$ means we apply different mixing ratio $\lambda$ for different patches, and each $\lambda$ is sampled from the original Beta distribution. Examples of these randomization will be shown in the appendix. From Table \ref{tab:randomness} we can conclude that not all randomness can lead to better performance. Except LUMix, both the random shuffling and random $\lambda$ lead to performance drop in terms of top-1 accuracy on ImageNet. Since random shuffling and random $\lambda$ both manipulate in the input space whereas LUMix is performed in the label space, adding randomness in the label space might be more promising for future works to explore.

\section{Conclusion}
In this paper, we observe that the label space for Mixup-based methods should consider some extent of ambiguity since it is hard even for the human beings to measure how uncertain they are when estimating the confidence for each image in an image mixture. To bridge the gap between the original pseudo-labeling and the human annotating process, we present LUMix that simply add two factors to simulate the human annotation. LUMix is embarrassingly simple and can be implemented within just a few lines of code. Extensive experiments demonstrate the efficacy of LUMix. Concretely, LUMix improves the top-1 accuracy of DeiT-S and XCiT-L by $0.7\%$ and $0.6\%$ respectively on ImageNet-1k for image classification. We also show that such superiority can be brought to the downstream tasks such as semantic segmentation. 

\paragraph{Limitation} 
LUMix is simple yet effective. We also demonstrated in the paper that it can lead to significant improvement for many different models. However, we admit that in this paper, LUMix is not backed by any theoretical proof. We did a lot of investigations to understand what stands behind, but none of them points to a clear conclusion. Due to its simplicity and efficacy, we think LUMix can attract considerable attention in the corresponding fields even without theoretical proof. We leave it to future works.

{\small
\bibliographystyle{ieee_fullname}
\bibliography{egbib.bib}
}

\end{document}